# Safety-oriented pedestrian trajectory prediction at urban intersections using time-to-collision and crossing-zone context

Erel Avineri[a], Yftach Gil[b], Yehudit Aperstein[b]

[a] Energy and Power Systems Engineering, Afeka Academic College of Engineering, Tel Aviv, Israel
[b] Intelligent Systems, Afeka Academic College of Engineering, Tel Aviv, Israel

## Abstract

Accurate pedestrian trajectory prediction is important for proactive road-safety applications, particularly at urban intersections where pedestrian motion is shaped by both vehicle interactions and crossing context. This study presents a safety-oriented trajectory-prediction framework that combines pedestrian motion history with Time-to-Collision (TTC) information and crossing-zone indicators. Using naturalistic trajectories from one urban intersection in the inD (Intersection Drone) dataset, several neural architectures were evaluated with 1.6 s observation and 2.4 s prediction horizons. A pooled Long Short-Term Memory (LSTM) separately encodes TTC histories and crossing-zone context before integrating them with pedestrian positions. In addition to conventional Average Displacement Error (ADE) and Final Displacement Error (FDE), prediction performance was assessed using the frequency and magnitude of errors exceeding a study-defined 1 m tolerance. A weighted loss was also introduced to place greater training emphasis on large coordinate-wise errors. Applying this loss to the position-only LSTM reduced ADE from 0.210 to 0.190 m and FDE from 0.550 to 0.503 m, while reducing ADE and FDE exceedance counts by 34.8% and 19.8%, respectively. The final pooled configuration incorporating TTC and crossing-zone information achieved an ADE of 0.184 m and FDE of 0.491 m, with further reductions of 33.5% and 6.3% in ADE and FDE exceedance counts relative to the safety-oriented position-only LSTM. The results indicate that safety-oriented training and structured integration of interaction and contextual information can reduce large trajectory-prediction errors, although broader validation across pedestrians, sites, and datasets is required.

## 1. Introduction

Pedestrians are among the most vulnerable road users in urban traffic systems. Understanding and analysing the risks pedestrians face when crossing roads has been the subject of extensive research addressing traffic conditions, road design, traffic control, and road-user behaviour (Ni et al., 2016; Orsini et al., 2023). Pedestrian detection has become an important computer-vision function in autonomous and assisted-driving systems, enabling vehicles to detect human movement in their path. Such detection is supported by integrated sensing technologies including radar, cameras, and light detection and ranging (LiDAR) (Iftikhar et al., 2022). These perception capabilities form an important component of Advanced Driver Assistance Systems (ADAS) and automated-driving systems aimed at improving road safety.

Despite substantial advances in vehicle safety technologies and infrastructure design, pedestrian fatalities and serious injuries remain a persistent challenge in cities worldwide, particularly at urban intersections where interactions between vehicles and pedestrians are frequent and complex (European Commission, 2022). In such environments, crashes are typically preceded by short-lived conflict situations, during which subtle and rapidly evolving behavioral decisions by pedestrians and drivers determine whether a collision occurs or is successfully avoided.

Quantitative safety data further underline the central role of pedestrian-vehicle interactions in urban crash risk. According to the European Road Safety Observatory, collisions involving motor vehicles account for the majority of pedestrian fatalities in urban areas. In 2022, car-related crashes were responsible for approximately 64% of pedestrian fatalities and 42% of cyclist fatalities in urban environments within the European Union (European Commission, 2022). These figures highlight the importance of interactions between vulnerable road users and motorized traffic for pedestrian safety in urban environments.

From a safety perspective, preventing pedestrian injuries requires not only detecting imminent collisions but also anticipating risky interactions sufficiently early to allow effective intervention. ADAS and automated driving functions increasingly rely on predictive components that estimate future pedestrian motion in order to support braking, warning, or evasive maneuvers. However, predicting pedestrian trajectories in real-world urban settings remains challenging due to the inherently adaptive and context-dependent nature of human behavior. Pedestrians continuously adjust their motion in response to approaching vehicles, surrounding traffic, and environmental cues, often in ways that are difficult to infer from position data alone.

Crash-based safety analysis is inherently retrospective and is limited by the relative rarity of severe events. Consequently, traffic safety research has long employed surrogate safety measures that quantify conflict severity before a crash occurs. Among these measures, Time-To-Collision (TTC) has been widely used as an indicator of imminent risk in pedestrian-vehicle interactions, as it reflects the remaining time until impact assuming constant motion of the interacting agents (Hayward, 1972; Vogel, 2003). Lower TTC values are generally associated with more critical pedestrian-vehicle interactions and have therefore been widely used to characterize interaction severity (Ni et al., 2016; Orsini et al., 2023).

Although TTC is well established as a surrogate safety measure, its use within pedestrian trajectory prediction remains comparatively limited. Existing approaches have incorporated TTC in different ways, including collision-risk-based interaction selection and encoding (Golchoubian et al., 2023b) and TTC-based interaction energy within recurrent pedestrian-interaction models (Dang et al., 2023). In the present study, TTC is instead represented as a continuous history of pedestrian–vehicle interaction values and jointly modeled with explicit crossing-zone context. This interaction and environmental representation is further combined with a safety-oriented loss and evaluation framework that explicitly emphasizes large trajectory-prediction errors. Most pedestrian trajectory prediction approaches focus on minimizing average spatial prediction error, typically evaluated using metrics such as Average Displacement Error (ADE) and Final Displacement Error (FDE). Although these metrics are useful for assessing general accuracy, they are not sufficient on their own for safety-oriented evaluation. In operational safety systems, rare but large prediction errors can be particularly consequential, as they may lead to delayed or incorrect interventions precisely in high-risk situations. Consequently, safety-oriented prediction models should be evaluated not only in terms of mean accuracy but also with respect to the occurrence and magnitude of large prediction errors.

In addition to interaction dynamics, pedestrian behavior at intersections is strongly influenced by environmental context. For example, crossing infrastructure, whether formally marked or informal, shapes pedestrian expectations (and the associated behavior) regarding vehicle yielding and acceptable risk. Empirical studies have shown that pedestrians behave differently in marked crosswalks compared to unmarked crossing locations, adjusting speed, hesitation, and gap acceptance accordingly (Ni et al., 2016). Capturing such contextual cues is therefore potentially important for understanding and predicting pedestrian motion in conflict-prone areas.

Motivated by these considerations, this study uses naturalistic urban-intersection trajectories to investigate safety-oriented pedestrian trajectory prediction. The analysis examines whether prediction can be improved by incorporating TTC-based surrogate safety information together with crossing-zone context. Specifically, we examine the integration of Time-to-Collision as a continuous interaction indicator between pedestrians and surrounding vehicles, together with crossing-zone indicators that encode thespatial context of pedestrian movement. Rather than focusing solely on average prediction accuracy, we emphasize the reduction of large prediction errors exceeding a defined tolerance threshold, which is used as a practical evaluation criterion rather than as a formally established safety standard.

The contributions of this study are fourfold.

- A safety-oriented evaluation framework that complements conventional ADE and FDE with threshold-exceedance measures quantifying both the frequency and magnitude of large trajectory-prediction errors is introduced. It extends evaluation beyond mean displacement accuracy toward explicit characterization of the upper-error tail.
- A threshold-weighted loss function is proposed to increase the training emphasis on large coordinate-wise prediction errors while retaining the standard quadratic treatment of smaller errors. The loss is designed to align model optimization with the proposed safety-oriented evaluation objective.
- A joint enriched representation that combines TTC information with multiple crossing-zone indicators is introduced, capturing complementary information on interaction urgency and infrastructure-related behavioral context.
- A structured pooled recurrent architecture is proposed to integrate pedestrian motion with TTC-based interaction information and crossing context. TTC and crossing-zone histories are encoded separately into compact latent representations before being fused with the pedestrian trajectory for recurrent prediction. This design provides a structured mechanism for integrating heterogeneous interaction and contextual signals and achieves the strongest performance among the evaluated configurations.

The rest of the paper is organized as follows. Section 2 reviews related work on pedestrian-vehicle conflicts, surrogate safety measures, and pedestrian trajectory prediction. Section 3 describes the dataset and study site. Section 4 presents the methodology, including feature construction, model design, and the safety-oriented loss and evaluation framework. Section 5 reports the experimental setup and results. Section 6 discusses the main findings and limitations. Section 7 concludes the paper and provides future research directions.

# 2. Background and Related Work

## 2.1 Pedestrian-Vehicle Conflicts and Safety-Oriented Analysis

Pedestrian safety in urban environments is commonly examined through the analysis of pedestrian-vehicle conflicts rather than crashes alone, as conflict situations provide richer information about risk mechanisms and behavioral adaptation preceding collision events. Pedestrian–vehicle conflicts at intersections can unfold over short time intervals, during which pedestrians and drivers continuously adjust speed, trajectory, and decision-making under uncertainty (Ni et al., 2016). Conflict-based approaches have therefore been widely adopted to evaluate safety interventions, infrastructure design, and behavioral patterns without relying exclusively on crash records (Orsini et al., 2023).

Such studies emphasize that pedestrian risk is not static but emerges dynamically from interaction timing, vehicle approach behavior, and pedestrian crossing decisions. This perspective motivates predictive approaches that seek to anticipate pedestrian motion before potentially hazardous interactions escalate.

## 2.2 Surrogate Safety Measures and Time-To-Collision

Surrogate safety measures are a cornerstone of proactive traffic safety research, enabling the assessment of interaction severity in the absence of collision events. Among these measures, Time-to-Collision (TTC) remains one of the most widely used indicators for quantifying imminent risk between road users. Initially introduced by Hayward (1972), TTC estimates the remaining time until a collision would occur under constant motion assumptions and has since been extensively applied in traffic safety studies (Vogel, 2003).

In pedestrian-vehicle contexts, TTC has been used to classify interaction severity, distinguish between safe and critical encounters, and analyze behavioral responses to approaching vehicles (Ni et al., 2016; Orsini et al., 2023). Its role in trajectory prediction has also received increasing attention. A systematic review by Golchoubian et al. (2023a) identified interaction modeling between pedestrians and vehicles as an important component of trajectory prediction in mixed-traffic environments. Golchoubian et al. (2023b) subsequently proposed the Polar Collision Grid approach, in which potentially interacting agents are selected according to collision risk and their interaction is encoded using TTC together with approach-direction information. The resulting polar representation focuses the prediction model on agents whose projected motion indicates a potential collision with the target pedestrian. Dang et al. (2023) introduced TTC-SLSTM, an extension of Social-LSTM that incorporates TTC-based interaction energy to represent collision-avoidance interactions between pedestrians. Rather than relying solely on spatial proximity, the interaction representation reflects the projected time until two pedestrians would collide if they maintained their current motion.

## 2.3 Predictive Modeling of Pedestrian Motion

Predicting pedestrian trajectories has become an important component of modern safety systems, including Advanced Driver Assistance Systems (ADAS) and automated driving platforms, such as autonomous vehicles (AVs). Anticipatory models allow vehicles to reason about future states of dynamic agents and plan interventions accordingly. Comprehensive surveys of human motion prediction highlight the role of trajectory forecasting in safe navigation within human-centered environments (Rudenko et al., 2020; Korbmacher and Tordeux, 2022).

A wide range of data-driven approaches has been proposed for pedestrian trajectory prediction, including recurrent neural networks, convolutional architectures, generative models, graph-based representations, and more recently transformer-based and large language model-inspired approaches (Alahi et al., 2016; Xue et al., 2018; Mangalam et al., 2020; Monti et al., 2021; Shi et al., 2021; Giuliari et al., 2021; Saleh, 2022; Alghodhaifi and Lakshmanan, 2023; Chib and Singh, 2025). These models have demonstrated increasing predictive accuracy across a variety of datasets and scenarios.

However, most predictive modeling studies evaluate performance primarily using average spatial error metrics, such as Average Displacement Error (ADE) and Final Displacement Error (FDE), which may obscure rare but large prediction failures. Zong et al. (2024) note that prediction performance often degrades in crowded or interaction-rich environments. Such settings are particularly challenging because pedestrian motion becomes more dependent on reactions to surrounding agents and therefore less predictable from past position alone.

## 2.4 Interaction Modeling and Environmental Context

Pedestrian motion is strongly influenced by interactions with surrounding agents and by environmental context. Interaction modeling techniques commonly encode the influence of neighboring entities - pedestrians or vehicles - through shared latent representations or pooling mechanisms, enabling models to capture social and spatial dependencies (Alahi et al., 2016; Xue et al., 2018; Monti et al., 2021). Such approaches acknowledge that pedestrian trajectories cannot be accurately predicted in isolation, particularly in mixed traffic environments and in heavily-crowded contexts.

Pedestrian crossing behavior also varies with individual characteristics; for example, age and fear of falling have been shown to influence behavior at crosswalks (Avineri et al., 2012). Environmental factors further shape pedestrian behavior, especially at urban intersections. Crossing infrastructure, road geometry, and spatial affordances influence pedestrian expectations regarding vehicle yielding and acceptable risk. Empirical studies have shown systematic behavioral differences between marked and unmarked crossings, affecting walking speed, hesitation, and gap acceptance (Ni et al., 2016). Saleh (2022), for example, incorporates scene semantics including roads, sidewalks, and zebra crossings into pedestrian trajectory prediction. Carrasco et al. (2021) instead models interactions among vehicles and vulnerable road users through a graph-based representation. The present study uses a more compact and directly interpretable environmental representation based on explicit crossing-zone indicators.

## 2.5 Datasets, Evaluation, and Safety-Relevant Limitations

Progress in pedestrian trajectory prediction has been supported by the availability of trajectory datasets captured under increasingly diverse real-world conditions. Datasets such as ETH (Pellegrini et al., 2009), UCY (Lerner et al., 2007), SDD (Robicquet et al., 2016), Waymo (Ettinger et al., 2021), RounD (Krajewski et al., 2020), and inD (Bock et al., 2020) have enabled systematic evaluation of predictive models under diverse conditions. These datasets span pedestrian-dominated environments, mixed road-user settings, and large-scale autonomous-driving scenarios. Naturalistic intersection datasets such as inD (Bock et al., 2020) are especially valuable for studying pedestrian-vehicle interactions in realistic urban traffic environments.

Recent studies have questioned whether conventional trajectory-prediction metrics adequately characterize model performance for safety-oriented applications. Rasouli (2020) shows that ADE and FDE remain the dominant measures for deterministic trajectory prediction, but also notes inconsistencies in their formulation across studies and emphasizes that probabilistic predictions require additional distribution-sensitive measures, such as negative log-likelihood or divergence-based metrics. Dendorfer (2023) raises a related concern for multimodal prediction: commonly used best-of-$N$ ADE/FDE measures evaluate only the prediction closest to the observed trajectory and can therefore overlook unrealistic alternative predictions generated by the same model. To address this limitation, Dendorfer advocates evaluating both the coverage and precision of the predicted trajectory distribution. Uhlemann et al. (2024) further demonstrate that suitability for autonomous driving cannot be inferred from displacement error alone, evaluating pedestrian predictors with respect to single-trajectory accuracy, dependence on observation history, inference time and scalability, as well as qualitative behavior in challenging motion scenarios. Together, these studies motivate evaluation frameworks that complement mean displacement accuracy with measures targeting application-relevant prediction behavior.

## 2.6 Summary and Research Gap

Existing research provides extensive insight into pedestrian-vehicle conflicts, surrogate safety measures, and predictive modeling of pedestrian motion. As reviewed above, TTC-derived interaction information has been incorporated into trajectory-prediction models (Golchoubian et al., 2023b; Dang et al., 2023), while other studies have incorporated environmental and crossing-related context (e.g., Saleh, 2022). However, these lines of research have largely developed separately, and limited attention has been given to their joint integration with evaluation and training mechanisms that explicitly target large prediction errors. This study brings these elements together by combining continuous TTC-based interaction histories with explicit crossing-zone context and a safety-oriented framework for training and evaluating large prediction errors on naturalistic urban-intersection data..

# 3. Data and Study Site

## 3.1 Naturalistic Trajectory Data for Safety Analysis

Empirical evaluation of pedestrian-vehicle interactions requires trajectory data that capture natural behavior under real-world conditions. Many commonly used pedestrian trajectory datasets focus on pedestrian-dominated or campus environments and therefore do not fully represent the mixed road-user interactions and infrastructure context encountered at urban intersections. As a result, findings derived from such datasets may not be generalized to common urban traffic environments involving frequent interactions between pedestrians and motorized vehicles.

To address these limitations, and in order to capture a typical urban environment and its pedestrian-car conflicts, this study uses the inD (Intersection Drone) dataset, a large-scale naturalistic trajectory dataset collected at urban intersections in Germany (Bock et al., 2020). Its mixed road-user composition provides the interaction environment required for the present analysis.

## 3.2 The inD Dataset

The inD dataset is a large-scale naturalistic trajectory dataset capturing road user behavior at urban intersections. Data were collected using camera-equipped drones flying at heights of up to 100 m, providing unobtrusive bird's-eye-view recordings that reduce occlusions while avoiding interference with natural road-user behavior (Bock et al., 2020).

The dataset comprises approximately 10 hours of traffic recordings from four unsignalized urban intersections in Aachen, Germany, all located on public roads with a speed limit of 50 km/h. It includes trajectories of 13,599 road users, covering pedestrians, cyclists, cars, buses, and trucks, with a substantial proportion of vulnerable road users. This diversity makes the dataset particularly suitable for analyzing pedestrian-vehicle interactions in typical urban settings.

Trajectories were extracted from 4K video recorded at 25 Hz using a vision-based tracking pipeline based on semantic segmentation, resulting in a reported positional accuracy of approximately 10 cm. Each road user is represented by a time-ordered trajectory consisting of spatial positions in a global ground-plane coordinate system, derived kinematic attributes and object-level metadata.

For each tracked road user, the available variables include the center position (x,y), heading, velocity and acceleration components, object class, and physical dimensions. In the present study, pedestrian position is used as the prediction target, whereas vehicle position, velocity, heading, and dimensions support the computation of interaction-related features described in Section 4.

## 3.3 Study Site Selection

Although the inD dataset contains recordings from four intersections, this study focuses on a single site, Location 2 (Frankenburg, Aachen). This location was selected based on its traffic composition, pedestrian density, and frequency of pedestrian-vehicle interactions. Location 2 features a four-arm unsignalized intersection situated near a residential area and a public park, resulting in a high volume of pedestrian and cyclist activity alongside regular motorized traffic. The site provides a substantial number of pedestrian trajectories together with frequent exposure to surrounding vehicle traffic. A summary of the number of trajectories by road user type and the total recording duration for this site is provided in Table 1. The site includes both a marked zebra crossing and several informal crossing areas, providing the spatial basis for the crossing-zone indicators introduced in Section 4.

**Table 1**. Summary of road-user trajectories and recording duration for Location 2 (Frankenburg, Aachen) in the inD dataset.

| Recording time [min] | Total tracks | Vehicles (cars, buses, trucks) | Bicycles | Pedestrians |
|---|---|---|---|---|
| 242 | 6235 | 2436 | 1700 | 2099 |

The spatial layout of the intersection is shown in Figure 1, illustrating the road geometry and pedestrian-crossing environment of the study site.

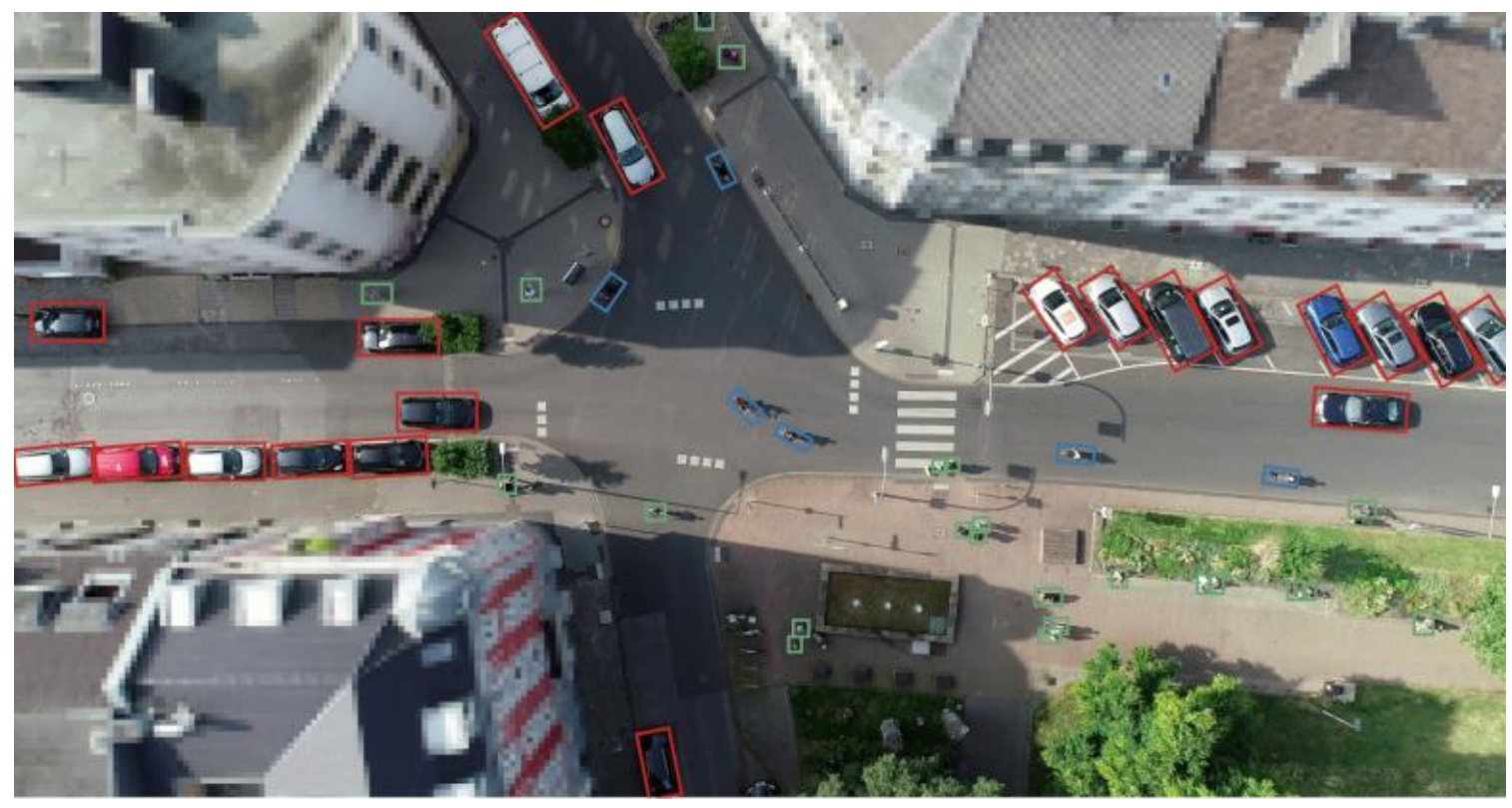

**Figure 1.** Aerial view of Location 2 (Frankenburg, Aachen), the four-arm unsignalized intersection used in this study.

Restricting the analysis to one intersection provides a controlled setting in which the relationship between pedestrian motion, vehicle interaction, and crossing context can be examined without introducing variability between different intersection layouts. At the same time, the single-site design limits the extent to which the results can be generalized to other road geometries and traffic environments, as discussed in Section 6.

### 3.4 Trajectory Representation and Temporal Resolution

Each road-user trajectory in the inD dataset is represented as a time-ordered sequence of spatial positions and derived kinematic attributes. In this study, pedestrian trajectories are modeled in two-dimensional ground-plane coordinates, while vehicle trajectories are used only to derive interaction-related features rather than as prediction targets. In addition to position, the dataset provides motion-related variables such as heading and velocity, which support the computation of interaction measures described in Section 4.

The original recordings were acquired at 25 frames per second. To balance behavioral fidelity with computational efficiency, the data were downsampled to 5 Hz, corresponding to one observation every 0.2 s. This sampling rate was used throughout the experiments and defines the temporal resolution of both the observation and prediction sequences. The generation of input-output trajectory samples from these downsampled tracks is described in Section 4.

## 4. Methodology

### 4.1 Problem Formulation: Predicting Pedestrian Motion under Conflict Risk

Pedestrian trajectory prediction is formulated here as a safety-oriented sequence-to-sequence forecasting task. Given an observed sequence of pedestrian states over a finite time window, the objective is to predict the pedestrian's future two-dimensional trajectory over a prediction horizon relevant to proactive safety intervention. In contrast to formulations based solely on motion history, the present task conditions prediction on both interaction-related information derived from surrounding vehicles and contextual information describing crossing infrastructure.

Let $\mathbf{p}_t = (x_t, y_t)$ denote the ground-plane position of pedestrian $i$ at discrete time step $t$, where time is indexed after temporal downsampling of the raw trajectories. For an observation window of length $T_{\text{obs}}$, the observed pedestrian trajectory is

$$\mathbf{P}^{\text{obs}} = \{\mathbf{p}_{t-T_{\text{obs}}+1}, \dots, \mathbf{p}_t\}.$$

The prediction target is the future trajectory over a horizon of length $T_{\text{pred}}$,

$$\mathbf{P}^{\text{pred}} = \{\mathbf{p}_{t+1}, \dots, \mathbf{p}_{t+T_{\text{pred}}}\}.$$

In addition to motion history, prediction is conditioned on an interaction feature sequence $\mathbf{T}^{\text{obs}}$ derived from TTC relationships with nearby vehicles, and on a contextual feature sequence $\mathbf{Z}^{\text{obs}}$ encoding whether the pedestrian is located within **each of four** predefined crossing zones. The predictive task is therefore defined as

$$\widehat{\mathbf{P}}^{\text{pred}} = f_{\theta}\left(\mathbf{P}^{\text{obs}}, \mathbf{T}^{\text{obs}}, \mathbf{Z}^{\text{obs}}\right),$$

where $f_{\theta}$ denotes the learned forecasting model. This formulation reflects the assumption that pedestrian motion at urban intersections is shaped not only by his/her past movement but also by evolving vehicle interaction risk and spatial crossing context.

An overview of the complete processing pipeline is provided in Figure 2. Pedestrian trajectories provide the motion history, surrounding-vehicle trajectories are used to derive TTC-based interaction features, and crossing-zone indicators provide spatial context; these inputs are integrated by the prediction model to estimate the future pedestrian trajectory.

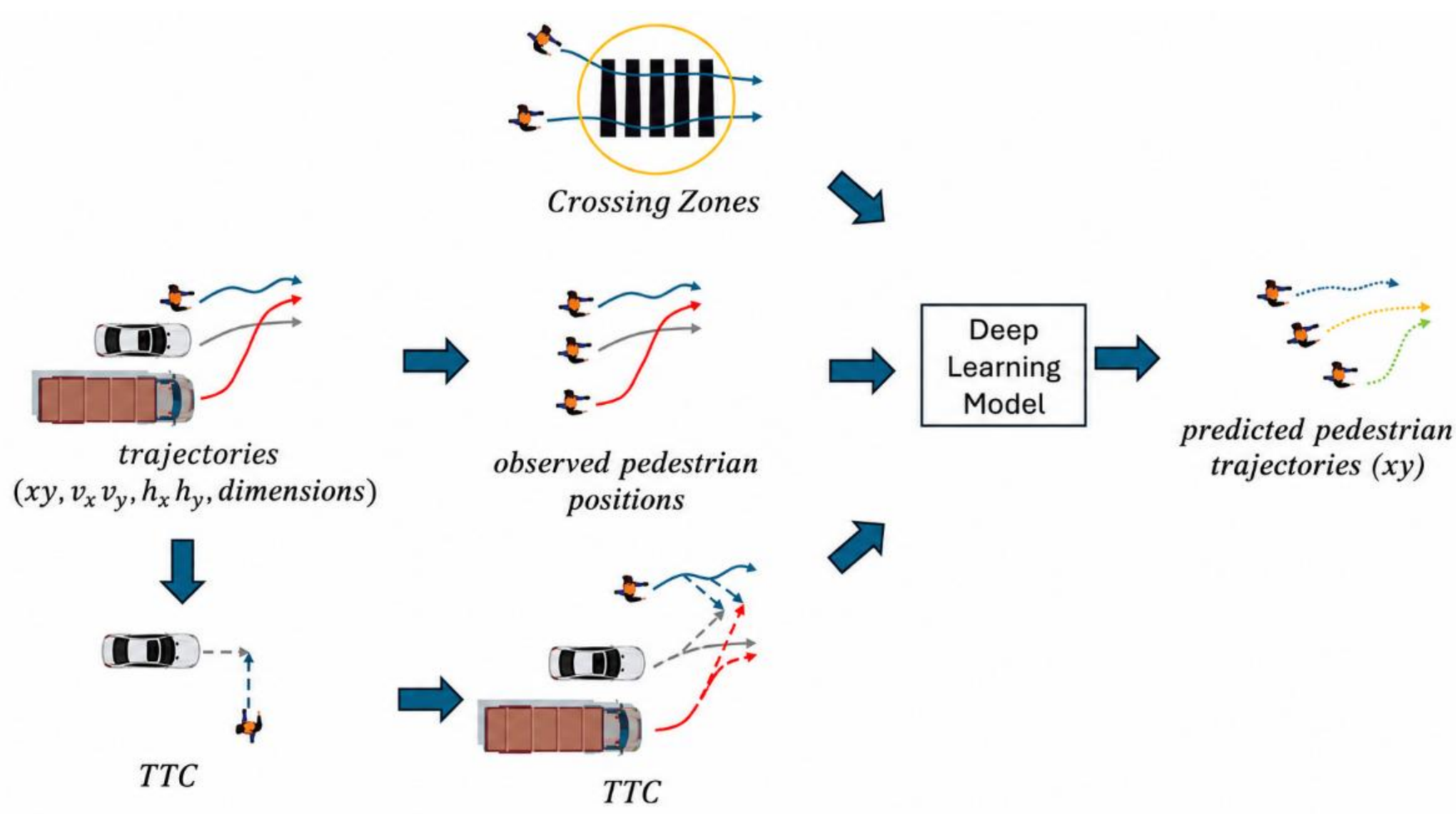

**Figure 2.** Overview of the safety-oriented pedestrian trajectory prediction framework. An observation sequence of pedestrian positions is combined with TTC-based interaction information and crossing-zone indicators to predict the future pedestrian trajectory.

## 4.2 Sample Generation and Prediction Protocol

Each eligible pedestrian track was segmented into fixed-length input-output sequences used for model training and evaluation. Each sequence consists of an observed trajectory segment, the associated interaction and contextual feature sequences over the observation window, and the subsequent pedestrian trajectory over the prediction horizon. This representation places all evaluated models within a common sequence-to-sequence prediction setting.

The observation window was set to 8 time steps and the prediction horizon to 12 time steps, corresponding to 1.6 s of observed motion and 2.4 s of future motion, respectively, at the downsampled trajectory rate of 5 Hz (0.2 s per time step). These values fall within the temporal ranges commonly used in pedestrian trajectory prediction and provide a short-term forecasting horizon that extends beyond immediate motion extrapolation while remaining relevant to near-term safety analysis. The 2.4 s horizon follows directly from the 12-step prediction setting at 5 Hz and represents a modeling choice, not a prescribed reaction-time or safety threshold.

A sample was generated from every pedestrian trajectory segment containing at least 20 consecutive downsampled observations. Pedestrian tracks shorter than this minimum duration were excluded, since they could not provide both a complete observation window and a complete prediction target. For each eligible pedestrian track, sequence samples were generated by sliding a fixed-length window along the trajectory timeline. The window was advanced by one downsampled time step, allowing multiple temporally overlapping samples to be generated from sufficiently long tracks.
For a window starting at time index $s$, the observed input trajectory is defined as

$$\mathbf{P}_s^{\mathrm{obs}} = \{\mathbf{p}_s, \dots, \mathbf{p}_{s+7}\},$$

and the corresponding prediction target is

$$\mathbf{P}_s^{\mathrm{pred}} = \{\mathbf{p}_{s+8}, \dots, \mathbf{p}_{s+19}\}.$$

For each observation time step, TTC-based interaction features were computed from the contemporaneous pedestrian-vehicle configuration, while crossing-zone indicators were derived from the pedestrian position relative to the manually defined crossing regions. Vehicle trajectories were therefore used only as auxiliary inputs for interaction modeling and were not themselves prediction targets.

## 4.3 Time-to-Collision Computation and Interpretation

TTC provides an intuitive and behaviorally meaningful measure of interaction urgency between a pedestrian and a surrounding vehicle. From a safety standpoint, TTC reflects how soon an interaction may become critical if both parties continue their current motion. Previous studies have shown that TTC is widely used to characterize the severity of pedestrian-vehicle interactions (Ni et al., 2016; Orsini et al., 2023).

TTC is computed under the constant-velocity assumption, which is commonly adopted in traffic safety analysis (Hayward, 1972; Vogel, 2003). For a pedestrian-vehicle pair $(i, j)$, TTC is defined as the remaining time until a collision would occur if relative motion and heading were to remain unchanged.

In the implemented procedure, TTC is evaluated in two dimensions using the spatial geometry of the pedestrian and vehicle bounding boxes rather than treating the road users as dimensionless points. The current positions, headings, velocities, and physical dimensions of the two road users are used to determine whether their continued motions would lead to an intersection of their bounding boxes and, if so, the corresponding positive collision time. For context, pedestrian-vehicle conflict studies have commonly treated TTC values below approximately 1.5 s as indicating severe or conflict-level interactions, values between about 1.5 and 3 s as more critical interactions, and values above 3 s as less critical, although the specific thresholds vary across studies and applications (Ni et al., 2016). In the present work, TTC is therefore retained as a continuous predictive input rather than converted into discrete risk categories.

In this study, TTC is computed at each time step between a pedestrian and multiple surrounding vehicles and incorporated as a time-varying interaction feature set within the predictive model. For each pedestrian, TTC values for up to nine surrounding vehicles are retained to form a fixed-size interaction representation. The choice of nine vehicle slots was based on an analysis of the study data, in which observations involving more than nine surrounding vehicles were rare. Thus, the interaction component at each observation time step consists of nine TTC entries. This design enables the model to represent the instantaneous interaction state between the pedestrian and surrounding vehicle traffic. The geometric interpretation of TTC computation is illustrated in Figure 3.

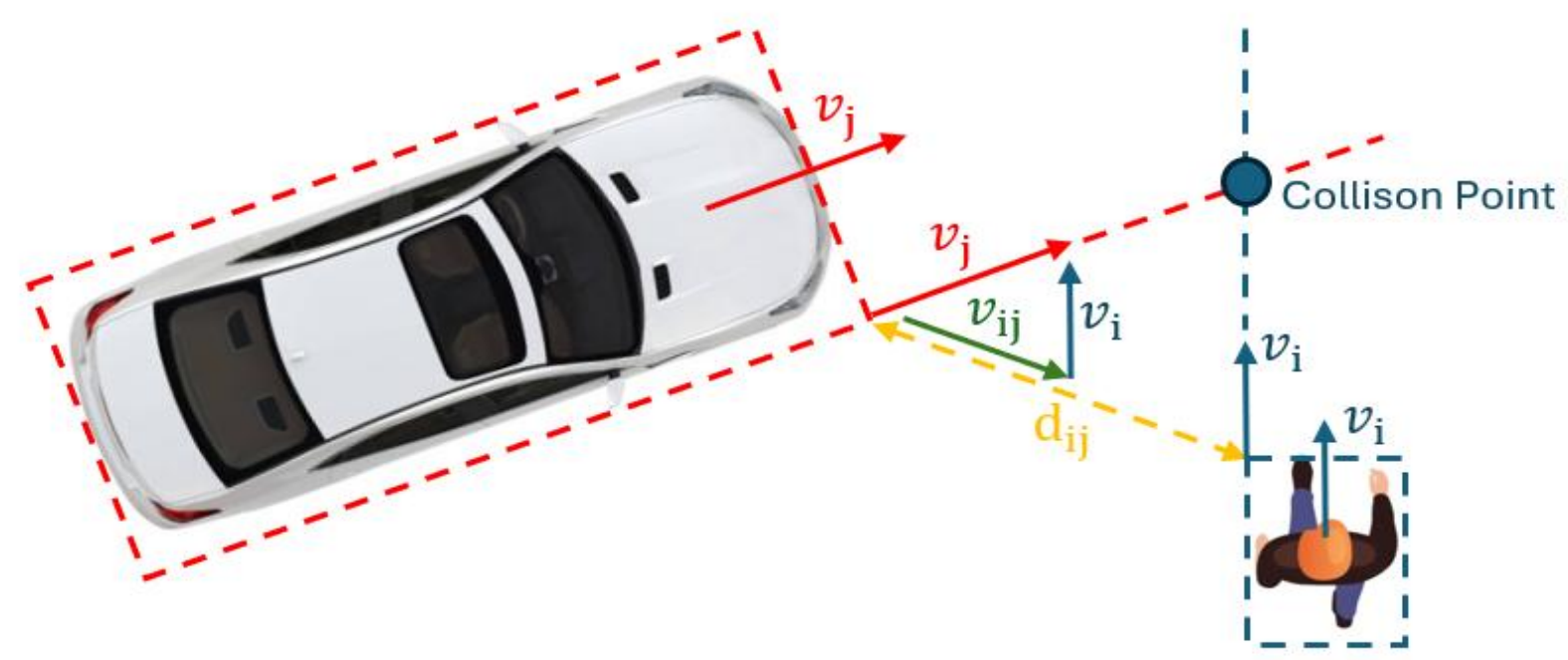

**Figure 3**. Conceptual illustration of the two-dimensional TTC calculation between a pedestrian and an approaching vehicle. Object dimensions, heading, and relative motion are taken into account through the road users' bounding-box geometry.

## 4.4 Crossing Zones as Environmental Context

Pedestrian behavior at urban intersections is influenced not only by dynamic interactions with vehicles but also by static infrastructural cues that shape expectations regarding right-of-way and acceptable risk. Marked crosswalks and informal crossing locations provide different behavioral affordances, affecting pedestrian speed, hesitation, and gap acceptance (Ni et al., 2016).
To incorporate this effect in a compact and interpretable manner, pedestrian crossing infrastructure is represented using crossing zones, defined as spatial regions associated with crossing behavior. These zones are manually defined for the study site based on intersection layout and observed pedestrian flow patterns.

Four crossing regions were represented: the marked zebra crossing ($Z_{\text{zebra}}$) and three informal crossing areas ($Z_1, Z_2, Z_3$). Each region is represented by a binary time-varying indicator. For zone $k$,

$$z_{k,t} = \begin{cases} 1, & \mathrm{p}_t \text{ lies within zone } k, \\ 0, & \text{otherwise.} \end{cases}$$

The contextual feature vector at time $t$ is therefore

$$\mathrm{z}_t = \left[z_{\text{zebra},t}, z_{1,t}, z_{2,t}, z_{3,t}\right] \in \{0,1\}^4.$$

Because the four crossing zones are mutually exclusive, the contextual vector $\boldsymbol{z}_t$is either the all-zero vector when the pedestrian is outside all defined crossing zones, or a one-hot vector with exactly one component equal to 1 when the pedestrian is located within one of the zones.

This representation allows the predictive model to condition motion estimates on infrastructural context without relying on dense semantic maps or image-based inputs, supporting a compact and directly interpretable representation of crossing location. The definition of crossing zones used in this study is illustrated in Figure 4.

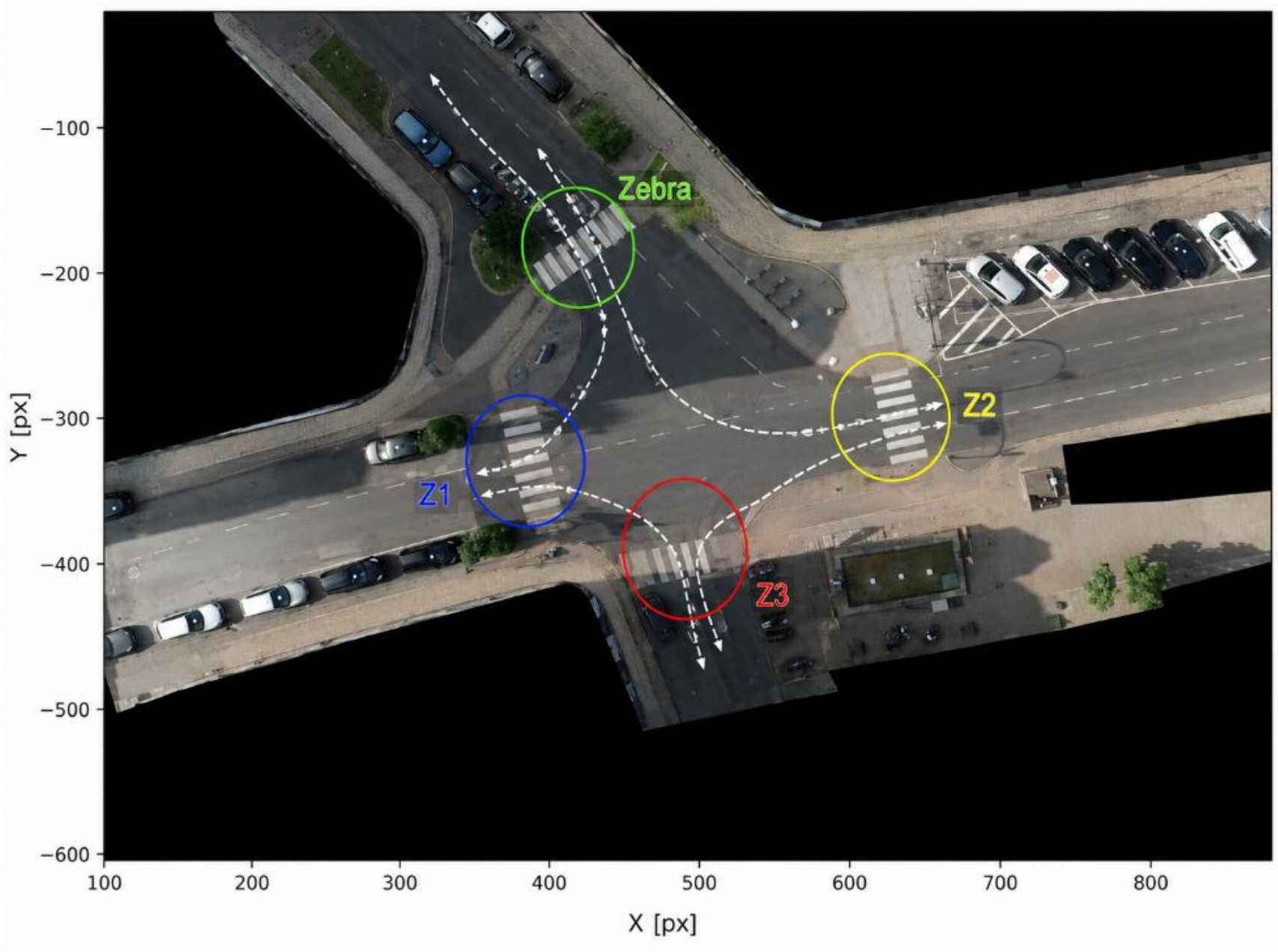

**Figure 4.** Four manually defined pedestrian crossing zones at Location 2: the marked zebra crossing (green) and three informal crossing areas Z1 (blue), Z2 (yellow), and Z3 (red). Each zone is encoded by a separate binary indicator according to whether the pedestrian is located within the corresponding region

## 4.5 Input Representation

At each observation time step, the input associated with pedestrian $i$ consists of three components: the pedestrian position vector $\mathbf{p}_t \in \mathbb{R}^2$, the TTC-based interaction feature vector $\boldsymbol{\tau}_t \in \mathbb{R}^9$, and the four-dimensional crossing-zone vector $\mathrm{z}_t \in \{0,1\}^4$.
The per-step enriched input vector is therefore defined as $\mathrm{x}_t = [\mathrm{p}_t, \tau_t, \mathrm{z}_t]$, with dimensionality $2 + 9 + 4 = 15$.

Over the full eight-step observation window, the enriched input sequence is therefore represented as $\mathrm{X}^{\mathrm{obs}} \in \mathbb{R}^{8\times15}$. The corresponding prediction target remains the future pedestrian trajectory, $\mathrm{Y} \in \mathbb{R}^{12\times2}$.

For the position-only baseline, the input is restricted to the pedestrian coordinates, thus providing a reduced input representation of dimension $8 \times 2$. This baseline enables comparison between position-only prediction and models incorporating TTC and crossing-zone information.

When fewer than nine surrounding vehicles are present at a given time step, the remaining TTC entries are padded to preserve the fixed nine-element interaction representation. In the preprocessing used for the experiments, unused TTC positions were represented by zero before model input, whereas non-collision TTC cases were mapped to a large finite sentinel value.

## 4.6 Compared Predictive Models

Four predictive models were considered in this study: a two-layer multilayer perceptron (MLP), an encoder-decoder (ED) architecture with latent noise injection, a standard Long Short-Term Memory (LSTM) network, and an LSTM with pooling. These models represent progressively richer approaches to pedestrian trajectory prediction than a simple feed-forward baseline to recurrent architectures capable of incorporating interaction-related and contextual information.

The MLP serves as the simplest baseline. It maps the fixed-length input representation directly to the predicted future trajectory without explicit temporal memory. The ED with latent noise provides a sequence-based baseline in which the observed input is encoded into a latent representation, perturbed with stochastic noise, and then decoded into the future trajectory. The standard LSTM serves as the main temporal baseline, processing the input sequence recurrently through hidden-state updates over the observation window. When enriched features are used, they are supplied directly to this model as part of the sequential input.

The LSTM with pooling is the proposed enriched-feature model. In this architecture, TTC-based interaction features and crossing-zone context are first encoded into dedicated latent representations before being fused with pedestrian motion history and passed to the recurrent predictor. The comparison therefore evaluates two different ways of incorporating the enriched features: direct concatenation in the standard LSTM and separate latent encoding in the pooled LSTM.

## 4.7 Proposed Pooled LSTM Architecture

The proposed model combines pedestrian motion history with encoded TTC-based interaction features and crossing-zone context. Importantly, the contextual and interaction encoders operate on the complete observation window rather than independently on each individual time step.

For an eight-step observation sequence, the nine TTC features are concatenated across time to obtain

$$\mathrm{T} \in \mathbb{R}^{8\times 9}, \qquad \mathrm{vec}(\mathbf{T}) \in \mathbb{R}^{72},$$

while the four crossing-zone indicators form

$$\mathrm{Z} \in \mathbb{R}^{8\times 4}, \qquad \mathrm{vec}(\mathbf{Z}) \in \mathbb{R}^{32}.$$

These two complete histories are separately mapped into low-dimensional latent representations:

$$\mathbf{h}_{\mathrm{TTC}} = g_{\mathrm{TTC}}\big(\mathrm{vec}(\mathbf{T})\big), \qquad \mathbf{h}_{\mathrm{zone}} = g_{\mathrm{zone}}\big(\mathrm{vec}(\mathbf{Z})\big).$$

In the implemented model, each encoder produces a two-dimensional latent representation. These representations are repeated across the eight observation time steps and concatenated with the corresponding pedestrian position, producing a six-dimensional recurrent input at each time step:

$$\mathrm{u}_t = [\mathbf{p}_t, \mathbf{h}_{\mathrm{TTC}}, \mathbf{h}_{\mathrm{zone}}] \in \mathbb{R}^6.$$

Over the observation window, the model processes the sequence $\{\mathrm{u}_t\}_{t=1}^{8}$ using an LSTM recurrent module. The pooled model used in the experiments contains three LSTM layers with a hidden-state dimension of 128.

Future positions are then generated autoregressively for 12 prediction steps. At each prediction step, the previously generated pedestrian position is used as the positional input, while the TTC and crossing-zone latent representations derived from the observation window remain fixed.

Compared with the standard LSTM baseline, the proposed architecture does not treat TTC and crossing-zone features only as appended scalar inputs. Instead, it encodes them into latent representations before recurrent prediction, enabling structured integration of interaction-related and contextual information. A schematic overview of the proposed pooled LSTM architecture is provided in Figure 5.

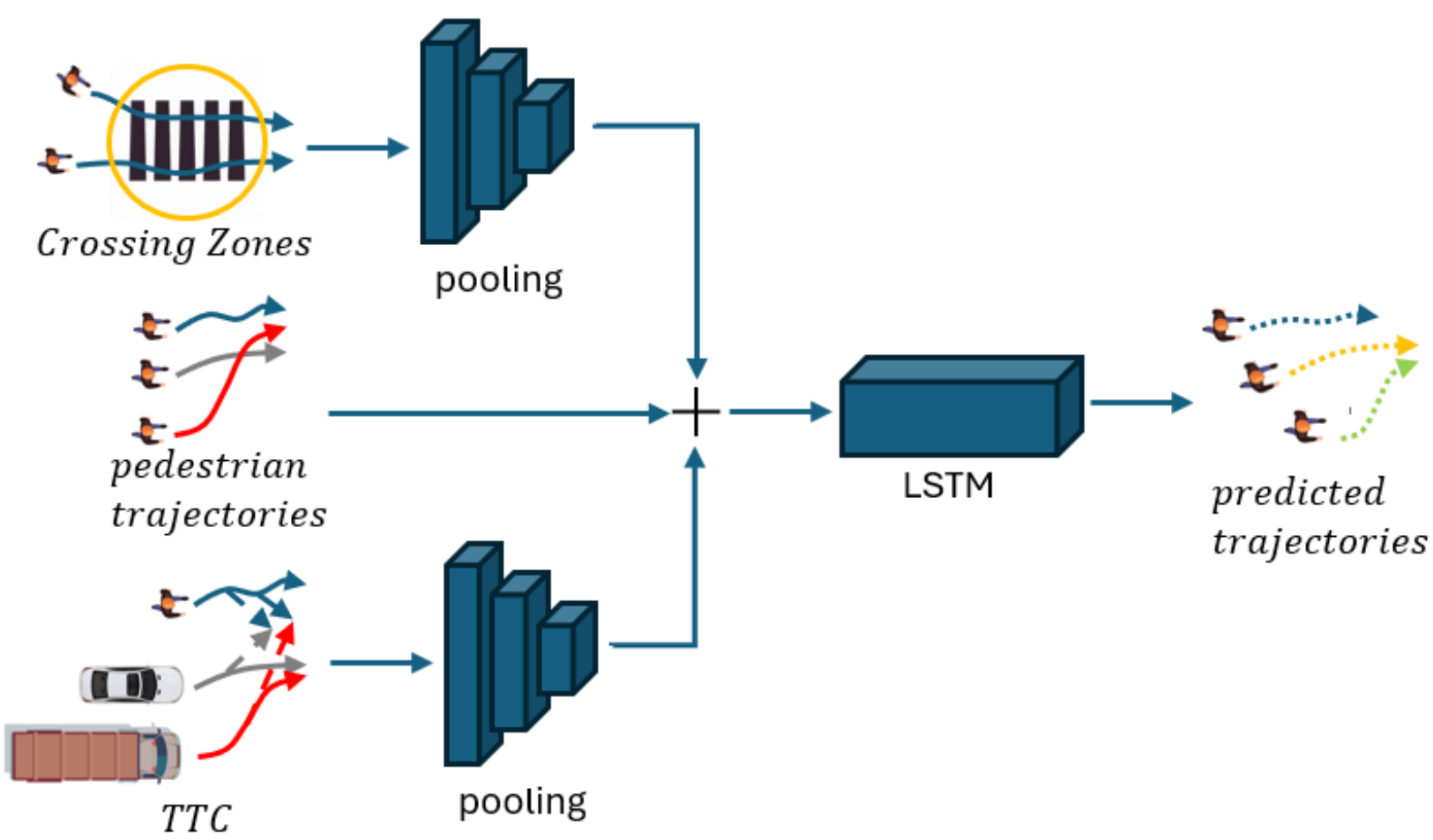

**Figure 5**. Pooled LSTM architecture for integrating pedestrian trajectories with TTC and crossing-zone context.

## 4.8 Safety-Oriented Loss Function

From a safety perspective, prediction errors are not equally consequential. Small deviations may be tolerable, whereas large errors can undermine timely intervention in high-risk situations. Accordingly, this study introduces a safety-oriented loss formulation that increases the training emphasis on large coordinate-wise prediction errors. The training objective is therefore based on a thresholded reweighting of the standard mean squared error.

For predicted future time step $t$ and coordinate $c \in \{x, y\}$, let the prediction error be

$$e_{t,c} = \hat{p}_{t,c} - p_{t,c}.$$

A binary penalty mask is defined coordinate-wise as

$$m_{t,c} = \begin{cases} 1, & |e_{t,c}| > P_t, \\ 0, & |e_{t,c}| \le P_t. \end{cases}$$

The corresponding weighted squared error is

$$\ell_{t,c} = e_{t,c}^2 [1 + (P_f - 1) m_{t,c}],$$

where $P_f > 1$ is the penalty factor. The training loss averages these weighted errors over all predicted time steps and both spatial coordinates are:

$$\mathcal{L} = \frac{1}{2T_{\text{pred}}} \sum_{t=1}^{T_{\text{pred}}} \sum_{c \in \{x,y\}} \ell_{t,c}.$$

Thus, prediction errors below $P_t$ retain their standard quadratic weight, whereas coordinate errors exceeding $P_t$ are upweighted by a factor of $P_f$.

## 4.9 Evaluation Metrics

Trajectory prediction quality is assessed using standard displacement-based metrics, complemented by threshold-exceedance measures to quantify the frequency and magnitude of large prediction failures at the sample level. Standard metrics facilitate comparability with prior work, while the additional measures characterize the upper-error portion of the prediction distribution rather than relying on mean performance alone.

Let $\mathrm{p}_{s,t}$ be the ground-truth pedestrian position and $\hat{\mathrm{p}}_{s,t}$ the predicted position for sample $s$ at future time step $t$. The pointwise Euclidean displacement error is

$$d_{s,t} = \|\hat{\mathrm{p}}_{s,t} - \mathrm{p}_{s,t}\|_2.$$

Average Displacement Error (ADE) measures the mean displacement error over the full prediction horizon:

$$ADE_s = \frac{1}{T_{\text{pred}}} \sum_{t=1}^{T_{\text{pred}}} d_{s,t}.$$

Final Displacement Error (FDE) measures the displacement error at the final predicted time step:

$$FDE_s = d_{s,T_{\text{pred}}}.$$

Dataset-level ADE and FDE are computed by averaging across all evaluated samples.

To capture rare but large prediction failures, an evaluation tolerance threshold $E_{\text{th}}$ is defined at the sample level. For each evaluated sample $s$, Threshold-exceedance measures are

$$I_s^{ADE} = \mathbb{1}(ADE_s > E_{\text{th_ADE}}), \qquad I_s^{FDE} = \mathbb{1}(FDE_s > E_{\text{th_FDE}}).$$

Dataset-level exceedance counts are then defined as

$$N_{ADE} = \sum_{s=1}^{S} I_s^{ADE}, \qquad N_{FDE} = \sum_{s=1}^{S} I_s^{FDE},$$

where $S$ is the number of evaluated samples.

To enable comparison across evaluation subsets, exceedance rates are also defined as

$$R_{ADE} = 100\frac{N_{ADE}}{S}, \qquad R_{FDE} = 100\frac{N_{FDE}}{S}.$$

In addition, for the subset of samples whose ADE or FDE exceeds $E_{\mathrm{th}}$, the mean and median exceedance values are reported in order to characterize the magnitude of large threshold-exceeding prediction errors. In the reported experiments, $E_{\mathrm{th_ADE}} = E_{\mathrm{th_FDE}} = 1$ m. This value is treated as a study-defined practical tolerance rather than as a standardized safety boundary.

# 5. Experiments and Results

## 5.1 Experimental Setup

The raw trajectories were first downsampled to 5 Hz and filtered to retain only pedestrian tracks of sufficient length (≥ 20 consecutive observations). The eligible pedestrian tracks were then randomly partitioned into an 80 % training subset and a 20 % test subset. Subsequent feature computation (TTC-based interaction features and crossing-zone indicators) and sequence generation were performed independently within each subset.

Sequence samples were formed with an observation window of $T_{\mathrm{obs}} = 8$ time steps and a prediction horizon of $T_{\mathrm{pred}} = 12$ time steps, corresponding to 1.6 s of observed motion and 2.4 s of future motion. The test subset contained 59,095 sequence samples generated from the held-out pedestrian tracks using the sliding-window procedure described in Section 4.2

The penalty parameters of the safety-oriented loss $(P_t, P_f)$ were examined using short pooled-LSTM training runs under different parameter combinations (Figure 6). Each combination was trained for 20 epochs, and Figure 6 reports the resulting number of samples whose ADE exceeded the 1 m evaluation threshold. Lower counts therefore indicate fewer large-error cases under the corresponding parameter setting. The setting $P_t = 1$ m and $P_f = 1.2$ yielded 355 ADE exceedances in this preliminary sensitivity analysis and was selected for the final experiments. The same setting was subsequently used for the final training runs of both the position-only LSTM and the pooled LSTM.

For the threshold-exceedance analysis, the evaluation tolerance was set to $E_{\mathrm{th}} = 1$ m, as defined in Section 4.9.

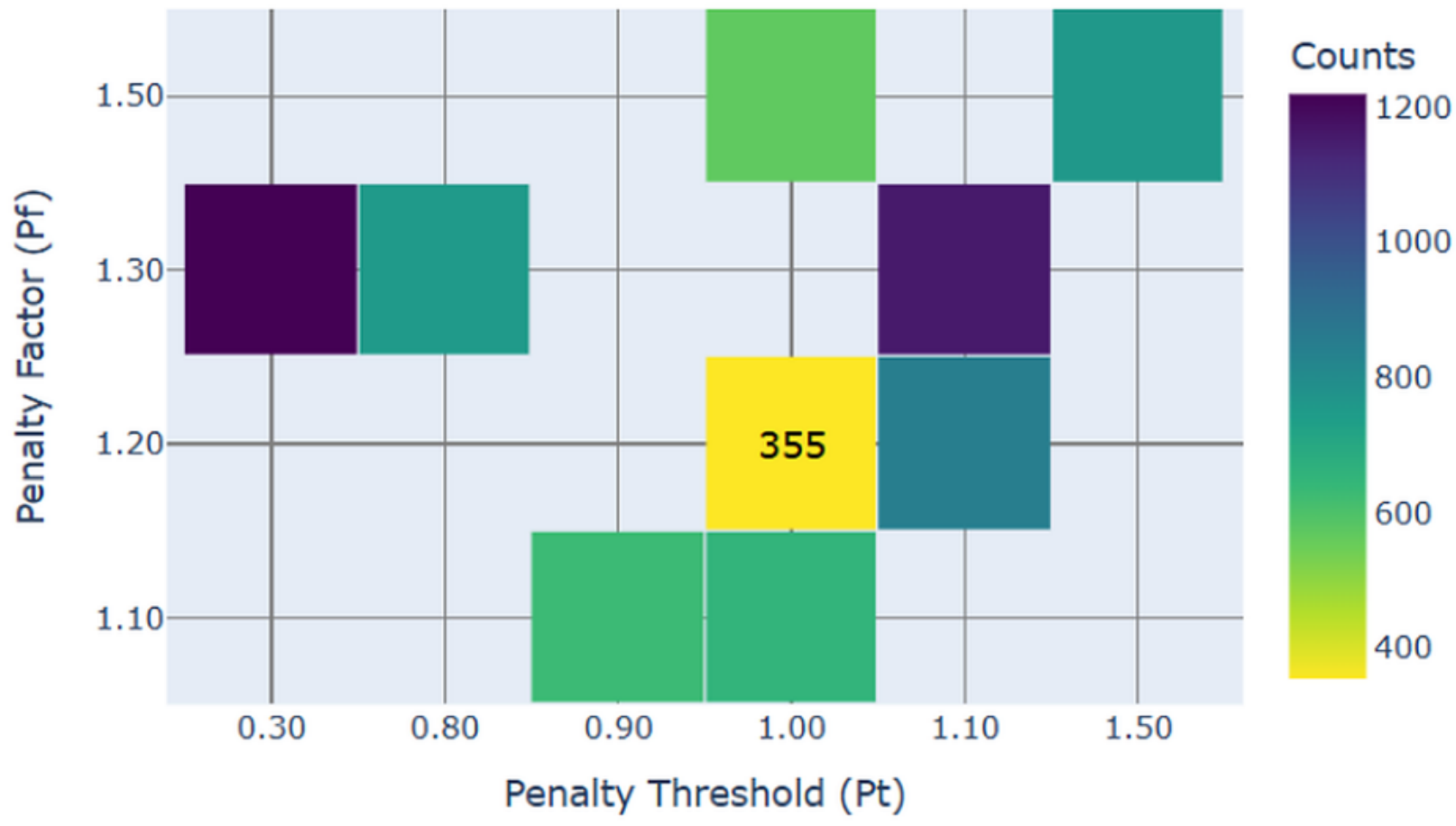

**Figure 6.** Sensitivity of the pooled LSTM to the safety-oriented loss parameters. The selected setting was $P_t = 1$ m and $P_f = 1.2$.

Models were trained in mini-batches of 64 using the Adam optimizer with a learning rate of $10^{-3}$, following the implementation used in the original experiments. The experiments were conducted in three stages. First, a baseline model-selection stage was performed under the position-only input setting using the standard loss. Second, TTC and crossing-zone features were added, and the comparison was extended to include the pooled LSTM to compare direct concatenation of the enriched features with their separate latent encoding. Third, the best position-only baseline and the best enriched-feature model were evaluated under the safety-oriented loss formulation using both conventional displacement metrics and threshold-exceedance measures.

## 5.2 Position-Only Baseline Comparison Under Standard Loss

The first experimental stage compared three predictive models under the position-only input setting: a two-layer MLP, an encoder-decoder with latent noise, and a standard LSTM. The purpose of this stage was to identify the best baseline predictor before introducing TTC and crossing-zone features. The results are reported in Table 2.

As shown in Table 2, the LSTM achieved substantially lower displacement errors than the other evaluated configurations and was therefore selected as the reference model for subsequent comparisons. The encoder-decoder with latent noise also performed competitively, whereas the two-layer MLP was markedly inferior in terms of displacement error. Within the set of models evaluated here, these results support the use of recurrent temporal modeling as the position-only baseline.

**Table 2.** Preliminary architecture comparison under the position-only input setting using the standard loss.

| **Model** | **Epochs** | **ADE [m]** | **FDE [m]** | **ADE count >1 m** | **FDE count >1 m** |
|---|---|---|---|---|---|
| 2-layer MLP | 40 | **25.859** | **89.595** | – | – |
| ED with noise | 500 | **0.266** | **0.703** | 811 | 11,572 |
| LSTM | 40 | **0.210** | **0.550** | 247 | 7,521 |

## 5.3 Enriched-Feature Comparison Under Standard Loss

In the second experimental stage, TTC and crossing-zone features were added to the input representation. Under this enriched setting, four models were compared: the MLP, the encoder-decoder with latent noise, the standard LSTM, and the pooled LSTM. This stage compared how the evaluated architectures handled the heterogeneous position, TTC, and crossing-zone inputs. Specifically, the standard LSTM receives the enriched features through direct concatenation, whereas the pooled LSTM first encodes TTC and crossing-zone features into dedicated latent representations before recurrent prediction. The corresponding results are presented in Table 3.

The results in Table 3 show that enriched features do not improve performance automatically. When TTC and crossing-zone features were supplied directly to the standard LSTM, performance deteriorated substantially. By contrast, the pooled LSTM achieved the best overall results under the enriched setting, showing that the evaluated pooled configuration handled the heterogeneous inputs more effectively than direct concatenation in the standard LSTM. The MLP again performed poorly, and the encoder-decoder with latent noise remained clearly inferior to the pooled LSTM.

**Table 3.** Preliminary architecture comparison under the enriched input setting using the standard loss.

| **Model** | **Epochs** | **ADE [m]** | **FDE [m]** | **ADE count >1 m** | **FDE count >1 m** |
|---|---|---|---|---|---|
| 2-layer MLP | 40 | **26.020** | **89.817** | – | – |
| ED with noise | 500 | **0.424** | **0.986** | 2,626 | 19,375 |
| LSTM | 40 | **1.562** | **3.732** | 32,431 | 52,718 |
| LSTM with pooling | 40 | **0.205** | **0.545** | 213 | 7,271 |

The results presented in Tables 2 and 3 support two observations. First, the LSTM provided the best position-only result among the evaluated baseline models. Second, TTC and crossing-zone features do not improve prediction automatically; under the evaluated configurations, direct concatenation substantially degraded LSTM performance, whereas the pooled configuration maintained low prediction error while incorporating the enriched information. These observations motivated the selection of the pooled LSTM as the main architecture for the final safety-oriented analysis. Because the standard and pooled LSTM configurations differ in their internal feature-processing architectures, these results should be interpreted as a comparison between the evaluated model configurations rather than as an isolated causal test of pooling alone.

## 5.4 Final Comparison and Threshold-Exceedance Analysis

After selecting the standard LSTM as the best position-only baseline and the pooled LSTM as the best enriched-feature model, the final comparison was performed under the safety-oriented loss formulation. In this stage, the standard LSTM was evaluated with position-only input, since this was its strongest-performing configuration, whereas the pooled LSTM was evaluated with the enriched input representation including TTC and crossing-zone features.

Before the final comparison, the penalty parameters of the pooled LSTM were tuned empirically in order to identify the most effective configuration for emphasizing large prediction errors during training. The parameter analysis identified Pt = 1 m and Pf = 1.2 m as the selected setting. The standard LSTM and pooled LSTM were then trained using this same safety-oriented loss configuration for the final comparison.

The final results are summarized in Table 4 for ADE-related measures and in Table 5 for FDE-related measures. To make the progression explicit, the tables include three configurations: the position-only LSTM trained with the standard loss, the same position-only LSTM trained with the safety-oriented loss, and the pooled LSTM trained with the safety-oriented loss using the enriched feature representation. Percent reductions shown in parentheses are calculated relative to the standard-loss position-only LSTM baseline.

Applying the safety-oriented loss to the position-only LSTM improved the baseline across both average-error and threshold-exceedance measures. Relative to the standard-loss baseline, ADE and FDE decreased by 9.5% and 8.5%, respectively, and the numbers of threshold-exceeding samples were reduced by 34.8% for ADE and 19.8% for FDE. When the pooled LSTM with TTC and crossing-zone features was evaluated under the same safety-oriented loss, further improvements were obtained over the safety-oriented position-only LSTM. Relative to the original standard-loss baseline, the final pooled configuration reduced ADE by 12.4%, FDE by 10.7%, the number of ADE exceedances by 56.7%, and the number of FDE exceedances by 24.8%.

**Table 4.** ADE-related comparison of the position-only baseline and final safety-oriented configurations. Percent reductions are relative to the standard-loss baseline.

| **Model** | **Input representation** | **ADE [m]** | **ADE count >1 m** | **ADE rate [%]** | **Mean exceed. [m]** | **Median exceed. [m]** |
|---|---|---|---|---|---|---|
| Baseline LSTM, standard loss | **Position only** | 0.210 | 247 | – | – | – |
| LSTM, safety-oriented loss | **Position only** | 0.190 (↓9.5%) | 161 (↓34.8%) | 0.272 | 1.229 | 1.143 |
| **Pooled LSTM, safety-oriented loss** | **Position + TTC + crossing zones** | 0.184 (↓12.4%) | 107 (↓56.7%) | 0.181 | 1.213 | 1.158 |

**Table 5.** FDE-related comparison of the position-only baseline and final safety-oriented configurations. Percent reductions are relative to the standard-loss baseline.

| **Model** | **Input representation** | **FDE [m]** | **FDE count >1 m** | **FDE rate [%]** | **Mean exceed. [m]** | **Median exceed. [m]** |
|---|---|---|---|---|---|---|
| Baseline LSTM, standard loss | **Position only** | 0.550 | 7,521 | – | – | – |
| LSTM, safety-oriented loss | **Position only** | 0.503 (↓8.5%) | 6,033 (↓19.8%) | 10.209 | 1.577 | 1.364 |
| **Pooled LSTM, safety-oriented loss** | **Position + TTC + crossing zones** | 0.491 (↓10.7%) | 5,655 (↓24.8%) | 9.569 | 1.536 | 1.341 |

Considering only the two models trained with the safety-oriented loss, the pooled configuration provided a further 3.2% reduction in ADE and 2.4% reduction in FDE relative to the position-only LSTM. More notably, the number of threshold-exceeding cases decreased by 33.5% for ADE and 6.3% for FDE. These incremental differences correspond to the comparison reported in the original project analysis.

The exceedance-magnitude statistics show a similar pattern. For both ADE and FDE, the pooled LSTM reduced the mean value of the threshold-exceeding subset relative to the safety-oriented position-only LSTM. For the median values, the pooled LSTM improved the FDE exceedance median, whereas the ADE exceedance median increased slightly. Thus, the final comparison shows a consistent reduction in most threshold-based error measures, with the clearest gains appearing in the frequency of threshold-exceeding ADE cases. Figure 7 further illustrates the distributions of these exceedances. Most ADE exceedances are concentrated relatively close to the 1 m threshold, whereas FDE exhibits a substantially longer right tail. The pooled configuration primarily reduces the frequency of threshold-exceeding cases rather than eliminating the occurrence of occasional large errors.

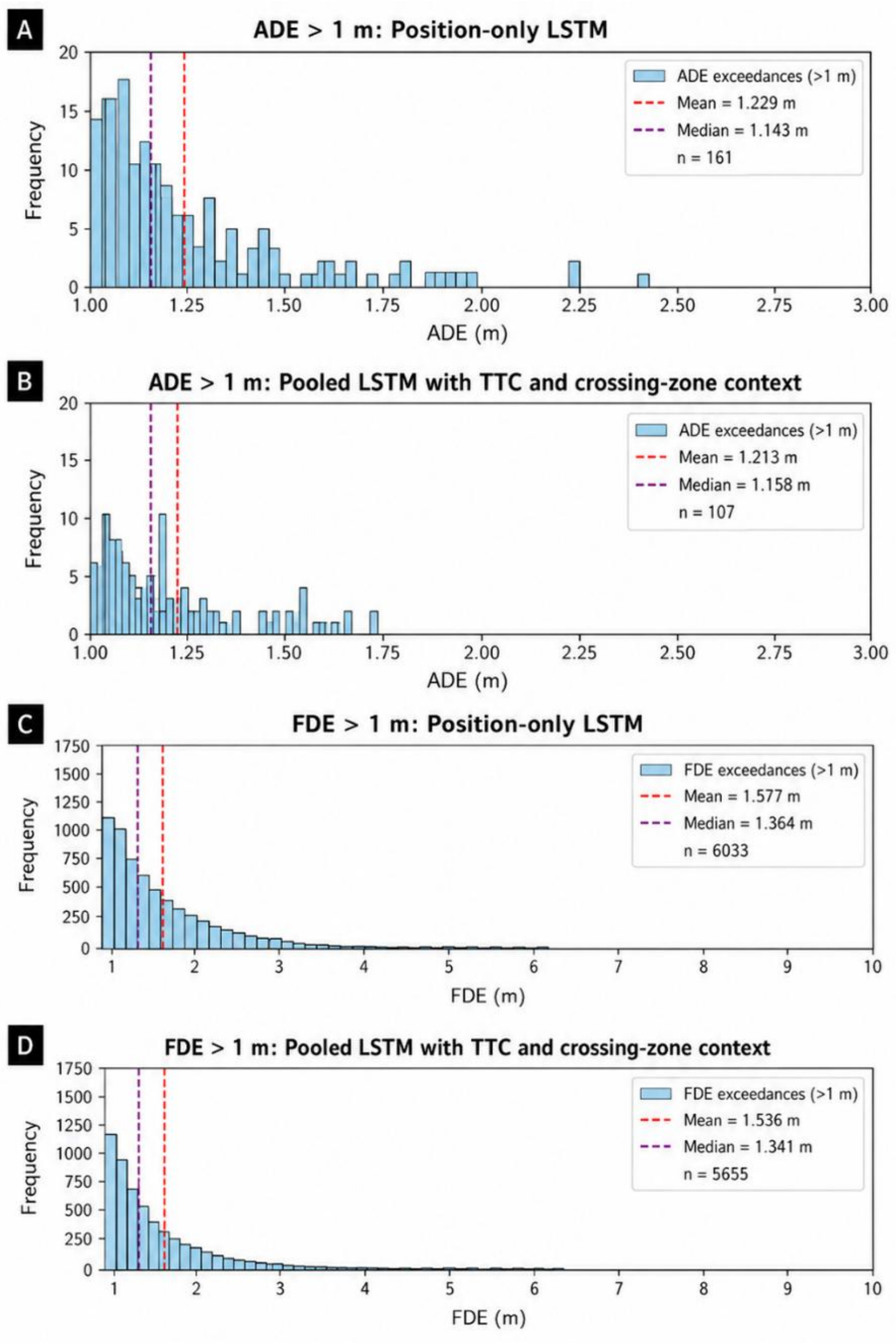

**Figure 7.** Distributions of threshold-exceeding ADE and FDE values for the safety-oriented position-only LSTM (A,C) and pooled LSTM configurations (B,D), including exceedance counts and distribution summaries.

# 6. Discussion

## 6.1 Main Findings and Safety-Oriented Interpretation

This study examined pedestrian trajectory prediction at an urban intersection from a safety-oriented perspective by combining conventional displacement accuracy with explicit analysis of large prediction errors. Three main observations emerge from the experiments.

First, introducing the safety-oriented loss improved the position-only LSTM across both conventional and threshold-based measures. Relative to the same LSTM trained with the standard loss, ADE decreased from 0.210 m to 0.190 m and FDE from 0.550 m to 0.503 m. More importantly for the proposed evaluation perspective, the number of samples exceeding the 1 m threshold decreased from 247 to 161 for ADE and from 7,521 to 6,033 for FDE. Thus, the weighted loss produced proportionally larger reductions in threshold-exceeding cases than in the corresponding mean displacement errors.

This distinction is relevant because mean prediction accuracy alone does not characterize the upper-error portion of the prediction distribution. A model may achieve a relatively small improvement in average displacement while substantially reducing the number of unusually inaccurate forecasts. In the present experiments, this pattern is particularly clear for ADE: the safety-oriented loss reduced mean ADE by 9.5%, whereas the number of ADE exceedances decreased by 34.8%. The results therefore support the use of complementary threshold-based measures when pedestrian trajectory prediction is evaluated from a safety-oriented perspective.

Second, the final pooled configuration incorporating TTC and crossing-zone context further improved performance relative to the position-only LSTM trained with the same safety-oriented loss. ADE and FDE decreased by approximately 3.2% and 2.4%, respectively, while the corresponding threshold-exceedance counts decreased by approximately 33.5% for ADE and 6.3% for FDE. The larger relative improvement in ADE exceedance frequency again indicates that differences between model configurations are more visible in the distribution of large errors than in mean displacement metrics alone. These findings are consistent with the motivation for considering both average accuracy and the frequency of large prediction failures.

The exceedance-magnitude statistics provide a more nuanced picture. The pooled model reduced the mean ADE and FDE among threshold-exceeding samples and reduced the median FDE exceedance. However, the median ADE exceedance increased slightly. Therefore, the enriched pooled configuration should not be interpreted as uniformly improving every characteristic of the error distribution. Its clearest advantage in the present experiments is the reduction in the number of threshold-exceeding predictions.

## 6.2 Role of TTC, Crossing Context, and Structured Feature Integration

The preliminary experiments show that simply adding TTC and crossing-zone variables to the input does not guarantee better prediction. Directly concatenating the enriched features with pedestrian coordinates in the standard LSTM resulted in substantially poorer performance, whereas the pooled LSTM retained prediction accuracy comparable to the position-only baseline and was subsequently improved through safety-oriented training. This suggests that heterogeneous interaction and contextual variables may require an appropriate representation before being combined with pedestrian motion history.

The pooled architecture addresses this issue by separately encoding the TTC history and crossing-zone history before combining these latent representations with pedestrian coordinates. Under the configurations evaluated here, this representation was more effective than direct feature concatenation. However, the experiments do not establish that pooling itself is universally necessary. The standard and pooled LSTM configurations differ not only in how the enriched features are processed but also in model capacity. Consequently, the observed differences cannot be attributed exclusively to the pooling operation.

Similarly, the final comparison does not isolate the independent contribution of TTC from that of crossing-zone context. Because TTC and crossing-zone features were introduced jointly, the present results characterize the combined enriched representation rather than the independent contribution of each feature family. Feature-specific analysis is therefore left for future work.

Nevertheless, the joint representation is conceptually relevant to pedestrian behavior at intersections. TTC supplies dynamic information about pedestrian-vehicle interaction under the assumed current motion, whereas crossing zones provide static spatial context related to where pedestrian movement occurs. These two sources of information therefore describe different aspects of the prediction problem: interaction state and environmental context.

## 6.3 Implications for Safety-Oriented Pedestrian Prediction

The results highlight a distinction between optimizing trajectory prediction for average accuracy and evaluating it for applications in which unusually large errors may be important. In an ADAS or automated-driving setting, trajectory forecasts can contribute to downstream functions such as pedestrian-risk assessment, warning, path planning, or braking decisions. Reducing the frequency of large forecasting errors could therefore be useful even when the improvement in mean ADE or FDE is modest.

At the same time, the present results should not be interpreted as evidence of direct crash-risk reduction. The study evaluates trajectory-prediction error rather than vehicle intervention performance or collision outcomes. Moreover, the 1 m evaluation threshold is a study-defined practical tolerance and not an empirically validated boundary separating safe from unsafe predictions. The threshold-based analysis should therefore be understood as a way of characterizing large prediction errors rather than as a direct measure of accident risk.

The TTC representation has a similar interpretation. TTC is a well-established surrogate measure of interaction severity, but in the present model it is used as predictive context rather than as an outcome measure. The resulting framework consequently links traffic-safety concepts with trajectory forecasting without claiming that prediction-error reductions translate directly into equivalent reductions in collision probability.

## 6.4 Limitations and Future Work

Several limitations should be considered when interpreting the findings. First, all experiments were conducted at a single inD intersection. The observed performance may therefore depend on the geometry, traffic composition, traffic volumes, and pedestrian behavior of this particular site, and external validity across other intersections remains to be established.

Second, the crossing zones were defined manually for the selected intersection. Although this provides a compact and interpretable representation of crossing context, it is site-specific and would require adaptation to different road layouts. Future work could derive such regions automatically from semantic maps, infrastructure annotations, or observed pedestrian flows. Future models might incorporate additional featues associated with the environmental context, such as road signs, traffic lights, pavement geometry and structure, etc.

Although it was not a scope of this work, future studies may incorporate features associated with the pedestrian characteristics, such as age, gender, disabilities, etc., as these might affect the pedestrian motion (see, for example, Avineri et al., 2012).

Finally, TTC relies on a constant-motion assumption and therefore provides only an instantaneous approximation of future interaction. Pedestrians and drivers continuously adapt their behavior to one another, whereas the current representation models vehicle influence on pedestrian prediction in a simplified form. Future models could incorporate bidirectional and multi-agent interaction mechanisms, including graph-based or attention-based architectures, and additional environmental information. The original project similarly identified multi-site validation and richer mutual interaction modeling as important extensions.

## 7. Conclusion

This study presents a safety-oriented framework for pedestrian trajectory prediction at urban intersections that combines pedestrian motion history with TTC-based pedestrian-vehicle interaction information and explicit crossing-zone context. In addition to conventional ADE and FDE metrics, the framework evaluates the frequency and magnitude of large prediction errors and introduces a weighted training loss that places greater emphasis on coordinate-wise errors exceeding a predefined threshold.

Within the evaluated inD intersection, applying the safety-oriented loss to the position-only LSTM reduced both average displacement errors and the number of threshold-exceeding predictions. The pooled LSTM incorporating TTC and crossing-zone information provided further improvements, reaching an ADE of 0.184 m and an FDE of 0.491 m. Relative to the position-only LSTM trained with the same safety-oriented loss, the pooled configuration reduced ADE and FDE by approximately 3.2% and 2.4%, while reducing ADE and FDE threshold-exceedance counts by approximately 33.5% and 6.3%, respectively.

These findings indicate that evaluating pedestrian trajectory prediction only through mean displacement error can obscure meaningful differences in the occurrence of large prediction failures. They also show that, under the evaluated configuration, TTC and crossing-zone information can be incorporated without degrading trajectory accuracy when processed through a structured latent representation.

Future work should evaluate pedestrian- and site-disjoint data, perform capacity-matched and feature-specific ablations, validate the approach on additional intersections and datasets, incoroporate additional features associated with the environmental context and the pedestrian characteristics, and investigate richer multi-agent interaction models. Such evaluation will be necessary to determine whether the observed reductions in large trajectory-prediction errors generalize to the broader range of conditions encountered by pedestrian-safety and automated-driving systems.

## Data and Code availability

The dataset and code are publicly available at:
https://github.com/yftach-gil/Pedestrian_Trajectory_Prediction_TTC_Crossing_Zones/tree/main

## References

Alahi, A., Goel, K., Ramanathan, V., Robicquet, A., Fei-Fei, L., & Savarese, S. (2016). Social LSTM: Human trajectory prediction in crowded spaces. In *Proceedings of the IEEE Conference on Computer Vision and Pattern Recognition (CVPR)* (pp. 961–971). https://doi.org/10.1109/CVPR.2016.110

Alghodhaifi, H., & Lakshmanan, S. (2023). Holistic spatio-temporal graph attention for trajectory prediction in vehicle–pedestrian interactions. *Sensors, 23*(17), 7361. https://doi.org/10.3390/s23177361

Avineri, E., Shinar, D., & Susilo, Y. O. (2012). Pedestrians' behaviour in crosswalks: The effects of fear of falling and age. *Accident Analysis & Prevention, 44*(1), 30–34. https://doi.org/10.1016/j.aap.2010.11.028

Bock, J., Krajewski, R., Moers, T., Runde, S., Vater, L., & Eckstein, L. (2020). The inD dataset: A drone dataset of naturalistic road user trajectories at German intersections. In *2020 IEEE Intelligent Vehicles Symposium (IV)* (pp. 1929–1934). IEEE. https://doi.org/10.1109/IV47402.2020.9304839

Carrasco, S., Llorca, D. F., & Sotelo, M. A. (2021). SCOUT: Socially-consistent and understandable graph attention network for trajectory prediction of vehicles and VRUs. In *2021 IEEE Intelligent Vehicles Symposium (IV)* (pp. 1501–1508). IEEE. https://doi.org/10.1109/IV48863.2021.9575874

Chib, P. S., & Singh, P. (2025). LG-Traj: LLM guided pedestrian trajectory prediction. In *Proceedings of the IEEE/CVF International Conference on Computer Vision Workshops (ICCVW)*. https://doi.org/10.1109/ICCVW69036.2025.00709

Dang, H.-T., Korbmacher, R., & Tordeux, A. (2023). TTC-SLSTM: Human trajectory prediction using time-to-collision interaction energy. In *2023 15th International Conference on Knowledge and Systems Engineering (KSE)*. IEEE. https://doi.org/10.1109/KSE59128.2023.10299443

Dendorfer, P. (2023). *Deep learning for human motion: Advancing trajectory prediction and multi-object tracking* [Doctoral dissertation, Technische Universität München].

Ettinger, S., Cheng, S., Caine, B., Liu, C., Zhao, H., Pradhan, S., Chai, Y., Sapp, B., Qi, C. R., Zhou, Y., Yang, Z., Chouard, A., Sun, P., Ngiam, J., Vasudevan, V., McCauley, A., Shlens, J., & Anguelov, D. (2021). Large scale interactive motion forecasting for autonomous driving: The Waymo Open Motion Dataset. In *Proceedings of the IEEE/CVF International Conference on Computer Vision (ICCV)* (pp. 9710–9719). https://doi.org/10.1109/ICCV48922.2021.00957

European Commission. (2022). *Annual statistical report on road safety in the EU*. European Road Safety Observatory.

Giuliari, F., Hasan, I., Cristani, M., & Galasso, F. (2021). Transformer networks for trajectory forecasting. In *2020 25th International Conference on Pattern Recognition (ICPR)* (pp. 10335–10342). IEEE. https://doi.org/10.1109/ICPR48806.2021.9412190

Golchoubian, M., Ghafurian, M., Dautenhahn, K., & Azad, N. L. (2023a). Pedestrian trajectory prediction in pedestrian-vehicle mixed environments: A systematic review. *IEEE Transactions on Intelligent Transportation Systems, 24*(11), 11544–11567. https://doi.org/10.1109/TITS.2023.3291196

Golchoubian, M., Ghafurian, M., Dautenhahn, K., & Azad, N. L. (2023b). Polar Collision Grids: Effective interaction modelling for pedestrian trajectory prediction in shared space using collision checks. In *2023 IEEE 26th International Conference on Intelligent Transportation Systems (ITSC)* (pp. 791–798). IEEE. https://doi.org/10.1109/ITSC57777.2023.10422509

Hayward, J. C. (1972). Near-miss determination through use of a scale of danger. *Highway Research Record, 384*, 24–34.

Iftikhar, S., Zhang, Z., Asim, M., Muthanna, A., Koucheryavy, A., & Abd El-Latif, A. A. (2022). Deep learning-based pedestrian detection in autonomous vehicles: Substantial issues and challenges. *Electronics, 11*(21), 3551. https://doi.org/10.3390/electronics11213551

Korbmacher, R., & Tordeux, A. (2022). Review of pedestrian trajectory prediction methods: Comparing deep learning and knowledge-based approaches. *IEEE Transactions on Intelligent Transportation Systems, 23*(12), 24126–24144. https://doi.org/10.1109/TITS.2022.3205676

Krajewski, R., Moers, T., Bock, J., Vater, L., & Eckstein, L. (2020). The rounD dataset: A drone dataset of road user trajectories at roundabouts in Germany. In *2020 IEEE 23rd International Conference on Intelligent Transportation Systems (ITSC)*. IEEE. https://doi.org/10.1109/ITSC45102.2020.9294728

Lerner, A., Chrysanthou, Y., & Lischinski, D. (2007). Crowds by example. *Computer Graphics Forum, 26*(3), 655–664. https://doi.org/10.1111/j.1467-8659.2007.01089.x

Mangalam, K., Girase, H., Agarwal, S., Lee, K.-H., Adeli, E., Malik, J., & Gaidon, A. (2020). It is not the journey but the destination: Endpoint conditioned trajectory prediction. In A. Vedaldi, H. Bischof, T. Brox, & J.-M. Frahm (Eds.), *Computer Vision – ECCV 2020* (pp. 759–776). Springer. https://doi.org/10.1007/978-3-030-58536-5_45

Monti, A., Bertugli, A., Calderara, S., & Cucchiara, R. (2021). DAG-Net: Double attentive graph neural network for trajectory forecasting. In *2020 25th International Conference on Pattern Recognition (ICPR)* (pp. 2551–2558). IEEE. https://doi.org/10.1109/ICPR48806.2021.9412114

Ni, Y., Wang, M., Sun, J., & Li, K. (2016). Evaluation of pedestrian safety at intersections: A theoretical framework based on pedestrian-vehicle interaction patterns. *Accident Analysis & Prevention, 96*, 118–129. https://doi.org/10.1016/j.aap.2016.07.030

Orsini, F., Batista, M., Friedrich, B., Gastaldi, M., & Rossi, R. (2023). Before-after safety analysis of a shared space implementation. *Case Studies on Transport Policy, 13*, 101021. https://doi.org/10.1016/j.cstp.2023.101021

Pellegrini, S., Ess, A., Schindler, K., & Van Gool, L. (2009). You'll never walk alone: Modeling social behavior for multi-target tracking. In *2009 IEEE 12th International Conference on Computer Vision* (pp. 261–268). IEEE. https://doi.org/10.1109/ICCV.2009.5459260

Rasouli, A. (2020). Deep learning for vision-based prediction: A survey. *arXiv*. https://doi.org/10.48550/arXiv.2007.00095

Robicquet, A., Sadeghian, A., Alahi, A., & Savarese, S. (2016). Learning social etiquette: Human trajectory understanding in crowded scenes. In B. Leibe, J. Matas, N. Sebe, & M. Welling (Eds.), *Computer Vision – ECCV 2016* (pp. 549–565). Springer. https://doi.org/10.1007/978-3-319-46484-8_33

Rudenko, A., Palmieri, L., Herman, M., Kitani, K. M., Gavrila, D. M., & Arras, K. O. (2020). Human motion trajectory prediction: A survey. *The International Journal of Robotics Research, 39*(8), 895–935. https://doi.org/10.1177/0278364920917446

Saleh, K. (2022). Pedestrian trajectory prediction for real-time autonomous systems via context-augmented transformer networks. *Sensors, 22*(19), 7495. https://doi.org/10.3390/s22197495

Shi, L., Wang, L., Long, C., Zhou, S., Zhou, M., Niu, Z., & Hua, G. (2021). SGCN: Sparse graph convolution network for pedestrian trajectory prediction. In *Proceedings of the IEEE/CVF Conference on Computer Vision and Pattern Recognition (CVPR)* (pp. 8994–9003). https://doi.org/10.1109/CVPR46437.2021.00888

Uhlemann, N., Fent, F., & Lienkamp, M. (2024). Evaluating pedestrian trajectory prediction methods with respect to autonomous driving. *IEEE Transactions on Intelligent Transportation Systems, 25*(10), 13937–13946. https://doi.org/10.1109/TITS.2024.3386195

Vogel, K. (2003). A comparison of headway and time to collision as safety indicators. *Accident Analysis & Prevention, 35*(3), 427–433. https://doi.org/10.1016/S0001-4575(02)00022-2

Xue, H., Huynh, D. Q., & Reynolds, M. (2018). SS-LSTM: A hierarchical LSTM model for pedestrian trajectory prediction. In *2018 IEEE Winter Conference on Applications of Computer Vision (WACV)* (pp. 1186–1194). IEEE. https://doi.org/10.1109/WACV.2018.00135

Zong, M., Chang, Y., Dang, Y., & Wang, K. (2024). Pedestrian trajectory prediction in crowded environments using social attention graph neural networks. *Applied Sciences, 14*(20), 9349. https://doi.org/10.3390/app14209349